\documentclass[10pt,twocolumn,letterpaper]{article}

\usepackage{cvpr}              

\usepackage[accsupp]{axessibility}  
\usepackage{graphicx}
\usepackage{amsmath}
\usepackage{amssymb}
\usepackage{booktabs}

\usepackage[pagebackref,breaklinks,colorlinks]{hyperref}

\usepackage[capitalize]{cleveref}
\crefname{section}{Sec.}{Secs.}
\Crefname{section}{Section}{Sections}
\Crefname{table}{Table}{Tables}
\crefname{table}{Tab.}{Tabs.}

\def\cvprPaperID{1343} 
\def\confName{CVPR}
\def\confYear{2023}

\usepackage{ragged2e}
\usepackage{booktabs}
\usepackage{color}
\usepackage{multirow}
\usepackage{bm}
\usepackage{bbm}

\makeatletter
\def\thanks#1{\protected@xdef\@thanks{\@thanks\protect\footnotetext{#1}}}
\makeatother

\begin{document}

\title{Learning with Fantasy: Semantic-Aware Virtual Contrastive Constraint for Few-Shot Class-Incremental Learning}

\author{Zeyin Song\textsuperscript{1}$^{\dag}$  \qquad Yifan Zhao\textsuperscript{2}$^{\dag}$ \thanks{$^{\dag}$: authors contributed equally. 
 $^{*}$: corresponding author.}\qquad Yujun Shi\textsuperscript{3} \qquad Peixi Peng\textsuperscript{2,4}$^{*}$\qquad Li Yuan\textsuperscript{1,4}
\qquad Yonghong Tian\textsuperscript{1,2,4}$^{*}$\\
\textsuperscript{1}School of Electronic and Computer Engineering, Peking University\\\
\textsuperscript{2}School of Computer Science, Peking University\\
\textsuperscript{3}National University of Singapore \qquad
\textsuperscript{4}Peng Cheng Laboratory\\
{\tt\small \textsuperscript{}zysong@stu.pku.edu.cn, \textsuperscript{}\{zhaoyf, pxpeng, yuanli-ece, yhtian\}@pku.edu.cn, \textsuperscript{}shi.yujun@u.nus.edu}}
\maketitle


\begin{abstract}

Few-shot class-incremental learning (FSCIL) aims at learning to classify new classes continually from limited samples without forgetting the old classes. The mainstream framework tackling FSCIL is first to adopt the cross-entropy (CE) loss for training at the base session, then freeze the feature extractor to adapt to new classes.
However, in this work, we find that the CE loss is not ideal for the base session training as it suffers poor class separation in terms of representations, which further degrades generalization to novel classes.
One tempting method to mitigate this problem is to apply an additional na\"ive supervised contrastive learning (SCL) in the base session. Unfortunately, we find that although SCL can create a slightly better representation separation among different base classes, it still struggles to separate base classes and new classes. Inspired by the observations made, we propose Semantic-Aware Virtual Contrastive model (SAVC), a novel method that facilitates separation between new classes and base classes by introducing \emph{virtual classes} to SCL. These virtual classes, which are generated via pre-defined transformations, not only act as placeholders for unseen classes in the representation space, but also provide diverse semantic information. By learning to recognize and contrast in the fantasy space fostered by virtual classes, our SAVC significantly boosts base class separation and novel class generalization, achieving new state-of-the-art performance on the three widely-used FSCIL benchmark datasets. Code is available at: \href{https://github.com/zysong0113/SAVC}{https://github.com/zysong0113/SAVC}.

\end{abstract}

\section{Introduction}
\label{sec:intro}

\begin{figure}
  \centering
  \begin{subfigure}{1\linewidth}
    \includegraphics[width=1\linewidth]{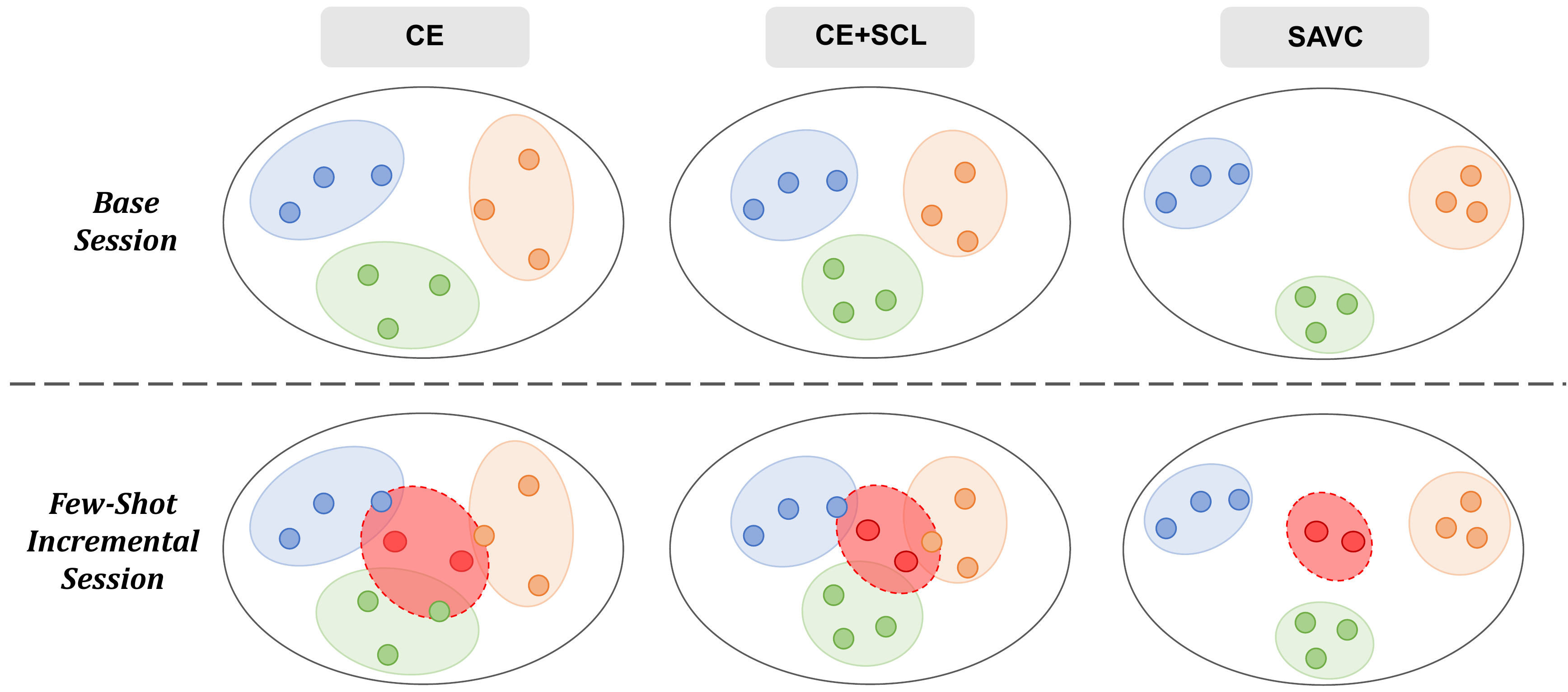}
    \caption{Comparison of the embedding spaces trained with different methods.}
    \label{fig:embed}
  \end{subfigure}
  \hfill
  \begin{subfigure}{1\linewidth}
    \includegraphics[width=1\linewidth]{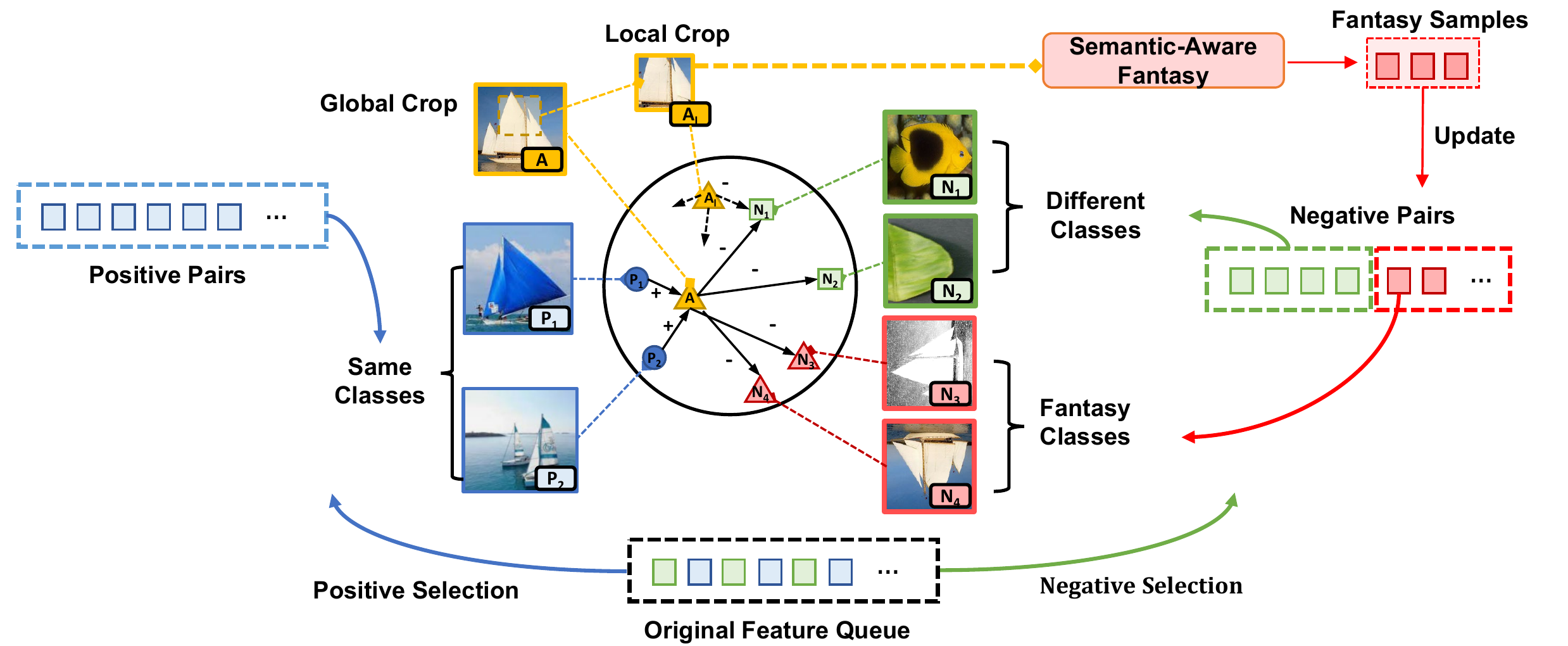}
    \caption{Illustration of our Semantic-Aware Virtual Contrastive model (SAVC).}
    \label{fig:illustrate}
  \end{subfigure}
  \caption{The motivation of the proposed approach. Under the incremental-frozen framework in FSCIL, our SAVC learns to recognize and contrast in the fantasy space, which leads to better base class separation and novel class generalization than CE and SCL.}
  \label{fig:motivation}
  \vspace{-10pt}
\end{figure}

Like humans, the modern artificial intelligence models are expected to be equipped with the ability of learning continually. For instance, a face recognition system needs to continually learn to recognize new faces without forgetting those has already kept in mind. In this case, Class-Incremental Learning (CIL)~\cite{masana2020class, hou2019learning,rebuffi2017icarl, yan2021dynamically, li2017learning} draws attention extensively, that sufficient and disjoint data are coming session-by-session. However, it is unrealistic to always rely on accessing of multitudinous data. Taking the case of face recognition system again, whether it can quickly learn a new concept with just a few images, is an important criterion for evaluating its performance. As a result, designing effective and efficient algorithms to resolve Few-Shot Class-Incremental Learning (FSCIL) problem has drawn increasing attention~\cite{dong2021few,tao2020few,chen2020incremental,shi2021overcoming,zhou2022forward,cheraghian2021semantic,zhang2021few}. Unlike CIL, only a few data are provided for FSCIL at incremental session while sufficient samples are only visible at the base session. 

The core challenge of CIL is to strike a balance of stability and plasticity, ~\ie, suppressing forgetting on old knowledge while adapting smoothly to new knowledge. Previous works on FSCIL have demonstrated that an incrementally trained model is prone to overfit on limited new data and suffer catastrophic forgetting~\cite{tao2020few,cheraghian2021semantic,dong2021few,zhao2021mgsvf}.
Thus the mainstream FSCIL framework~\cite{zhu2021self,shi2021overcoming,zhang2021few} is to first use the cross-entropy (CE) loss for training in the base session, then freeze the backbone to adapt to new classes. Given such baseline framework, the main problem that needs to be taken into consideration is:
\emph{What makes a good base-session-model that can generalize well to new classes with limited data?}

Intuitively, if all representations of base classes concentrate around their own clustering centers (\ie, prototypes), and all prototypes are far away from each other, then novel classes should be easily incorporated into the current space without overlapping.
Therefore, we conjecture that base class separation facilitates novel class generalization.
However, the current widely adopted CE loss in FSCIL cannot well separate the class margins, which causes poor generalization to new classes. This motivates our attempt at methods with better clustering effects, such as supervised contrastive learning (SCL)~\cite{khosla2020supervised}.
We conduct experiments to validate our conjecture by training models optimized by only CE loss, and both CE and SCL. We empirically find that CE leads to lower base class separation and classification accuracy than SCL. Although SCL indeed slightly improves the base class separation with higher accuracy, we find it implicitly decreases the inter-class distance. Therefore, the two methods both leave inadequate room for future updates, which leads to the inevitable overlapping between novel classes and old classes in the representation space (as shown in the left and middle columns of \cref{fig:embed}).

The unsatisfactory performance of SCL implies that the limited semantic information base session provided may need our fantasy remedy. Although we have no access to future classes in the base session, we can create many virtual classes with various semantics filling the unallocated representation space, which not only holds enough space for future classes, but also enables the learning of diverse semantics for better inference and generalization. 

To this end, we propose a Semantic-Aware Virtual Contrastive (SAVC) model to improve the FSCIL by boosting base class separation (see \cref{fig:illustrate}). Our fantasy space is constructed with pre-defined transformations~\cite{zhang2016colorful,pathak2016context,doersch2015unsupervised,gidaris2018unsupervised}, which can be regarded as the semantic extension of the original space in finer grains. The type of transformations represents the richness of our imagination,~\ie, how compact we want the embedding space to be and how fine the semantic grains we want to acquire. Learning to contrast and recognize in the fantasy space not only brings a better class separation effect, but also provides richer semantic information to improve the accuracy. In contrast to the CE baseline and SCL, our SAVC can simultaneously decrease intra-class distance and increase inter-class distance, which leads to better base class separation and novel class generalization (as given in the right of \cref{fig:embed}). Extensive experiments on various benchmark datasets further verify the superiority of our method. 

Our main contributions are summarized in three folds:

1) We empirically discover that it is crucial to boost class separation degree in base session for the incremental-frozen framework in FSCIL, which helps fast generalization for novel classes with only a few samples.

2) We propose a novel Semantic-Aware Virtual Contrastive (SAVC) framework, enhancing the FSCIL learner by learning to recognize and contrast in the fantasy space, which preserves enough place for novel generalization and enables a multi-semantic aggregated inference effect. 

3) Extensive experiments on three FSCIL benchmarks,~\ie, CIFAR100, \emph{mini}ImageNet and CUB200, demonstrate that our SAVC outperforms all approaches by a large margin and achieves the state-of-the-art performance.

\section{Related Work}
\label{sec:relatedwork}

\textbf{Few-Shot Learning.} FSL focuses on fast adaptation to unseen classes through few learnable samples. Current FSL methods are mainly divided into two lines: optimization-based methods ~\cite{li2017meta, ravi2017optimization,finn2017model,nichol2018first} and metric-based methods~\cite{triantafillou2017few,oreshkin2018tadam,snell2017prototypical,sung2018learning}. Optimization-based methods aim to encourage the model to obtain generalizable representations by using meta-learning frameworks. Our method has more connection with metric-based methods, which use a pre-trained backbone to extract feature, and compute score by measuring the distance between query and support. Nevertheless, FSL aims to adapt to novel classes via few samples and cannot be directly applied to FSCIL, without maintaining classification performance over base classes.

\textbf{Class Incremental Learning.} CIL is the process of learning new knowledge continually without forgetting. Classical works fall into three streams. The first stream rehearsals former samples selectively to maintain discrimination over old classes~\cite{rebuffi2017icarl,rolnick2019experience,chaudhry2018efficient}. The second stream regularizes the important parameters from being dramatically changed~\cite{li2017learning,liu2018rotate}. 
The third stream dynamically expands the network architecture to adjust the model’s representation ability on new tasks~\cite{yan2021dynamically, wang2022learning}.
Recent methods boost performance by encouraging the model to learn to recognize more classes~\cite{zhu2021class,shi2022mimicking}, which are relative to our method. CIL methods build on the basis of the large number of training examples, which is unsuitable for FSCIL.

\textbf{Few-Shot Class-Incremental Learning.} FSCIL~\cite{hersche2022constrained,cheraghian2021synthesized,liu2022few,chi2022metafscil,achituve2021gp,chen2020incremental,mazumder2021few, ahmad2022few} aims at undertaking CIL task in insufficient data settings. TOPIC~\cite{tao2020few} first proposes this new task and solves it by neural gas structure. Current FSCIL methods can be roughly divided into two groups. The first group updates the backbone in incremental session~\cite{tao2020few,cheraghian2021semantic,dong2021few,zhao2021mgsvf}, while the second group fixes the backbone in incremental sessions~\cite{zhu2021self,shi2021overcoming,zhang2021few}. Built on the latter incremental-frozen framework, other works balance the stability and plasticity trade-off
from different perspectives, such as finding flatten minima~\cite{shi2021overcoming}, episodic training~\cite{zhang2021few,zhu2021self,zhou2022few}, ensemble learning~\cite{ji2023memorizing}, forward compability~\cite{zhou2022forward}, and so on.

\textbf{Self-supervised learning.} SSL aims at learning representations without human annotations. A common way of SSL is to design pretext tasks and teach the model how to solve these prediction tasks~\cite{zhang2016colorful,pathak2016context,doersch2015unsupervised,gidaris2018unsupervised}.~\cite{lee2020self} proposes to utilize these transformations under full-supervised datasets and trains a unified classifier, which motivates the construction of our fantasy space. 
Another commonly used pretext task is instance discrimination represented by contrastive learning~\cite{he2020momentum,chen2020improved,caron2020unsupervised}. The methodology is to teach the model to aggregate positives and exclude negatives. SupCon~\cite{khosla2020supervised} converted to ``class discrimination" task, leveraging the label information to decide positive pairs. 


\section{Methodology}
\label{sec:method}

\subsection{Preliminary}
\label{sec:formulate}
In FSCIL, we learn a sequence of tasks continually. There is a constant stream of training data denoting $\mathcal{D}_{train} = \left\{ \mathcal{D}_{train}^{t} \right\}_{t = 0}^{T}$, where $\mathcal{D}_{train}^{t} = \left\{ \left( {\bm{{\rm x}}_{i},~y_{i}} \right) \right\}_{i = 0}^{N_{t}}$ represents the training samples from session $t$, $\bm{{\rm x}}_{i}$ and $y_{i}$ are the $i$-th image and corresponding label respectively. Assuming that the label space of the $t$-th task is $\mathcal{C}^{t}$, the label spaces from different tasks have no overlap,~\ie, the model trained on the $\mathcal{D}_{train}^{t}$ should be evaluated on $\mathcal{D}_{test}^{t}$ which contains all encountered classes $\mathcal{C}^{0} \cup \mathcal{C}^{1}\cdots \cup \mathcal{C}^{t}$ in the $t$-th session. Specifically, sufficient data can be provided in base session (also called initial session), while the model only has access to a few samples in all following sessions (also called incremental sessions). In particular, the incremental data are always organized as $N$-way $K$-shot format,~\ie, there are $N$ classes, and each class contains $K$ training images.

\begin{figure}
  \centering
  \begin{subfigure}{0.49\linewidth}
    \includegraphics[width=1.09\linewidth]{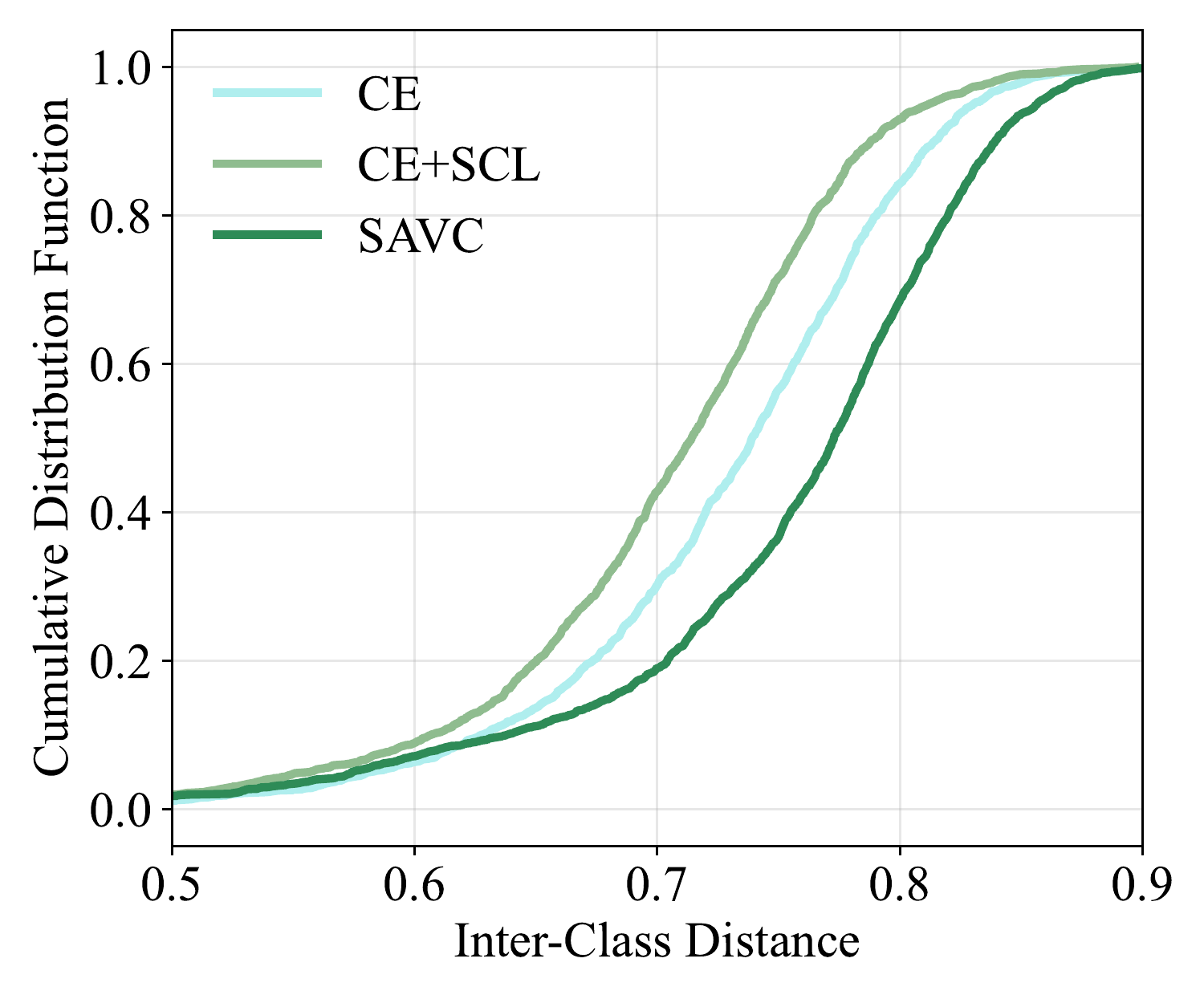}
    \caption{Cumulative distribution function of inter-class 
    distance.}
    \label{fig:inter}
  \end{subfigure}
  \hfill
  \begin{subfigure}{0.49\linewidth}
    \includegraphics[width=1.09\linewidth]{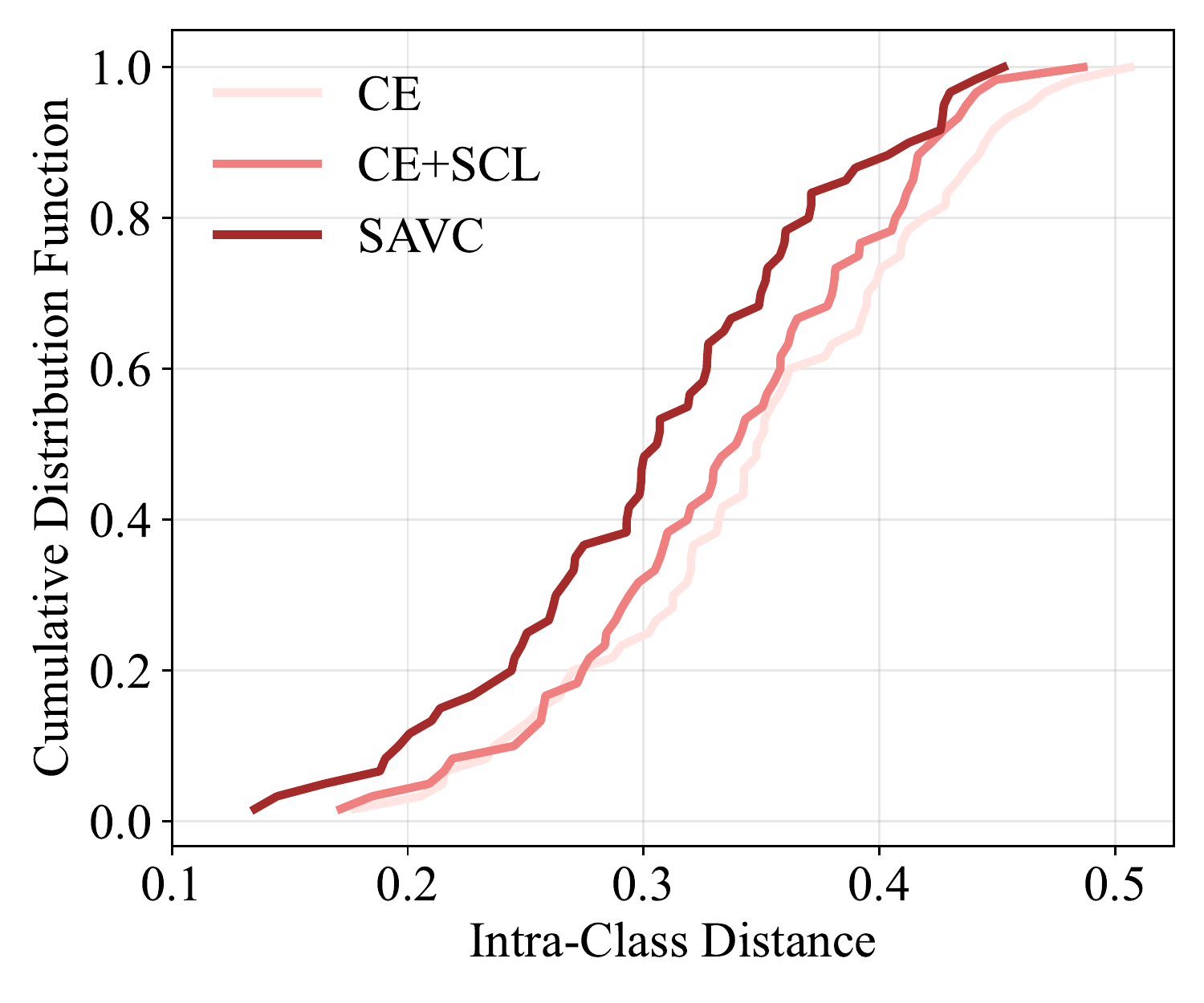}
    \caption{Cumulative distribution function of intra-class distance.}
    \label{fig:intra}
  \end{subfigure}
  \hfill
  \begin{subfigure}{0.49\linewidth}
    \includegraphics[width=1.05\linewidth]{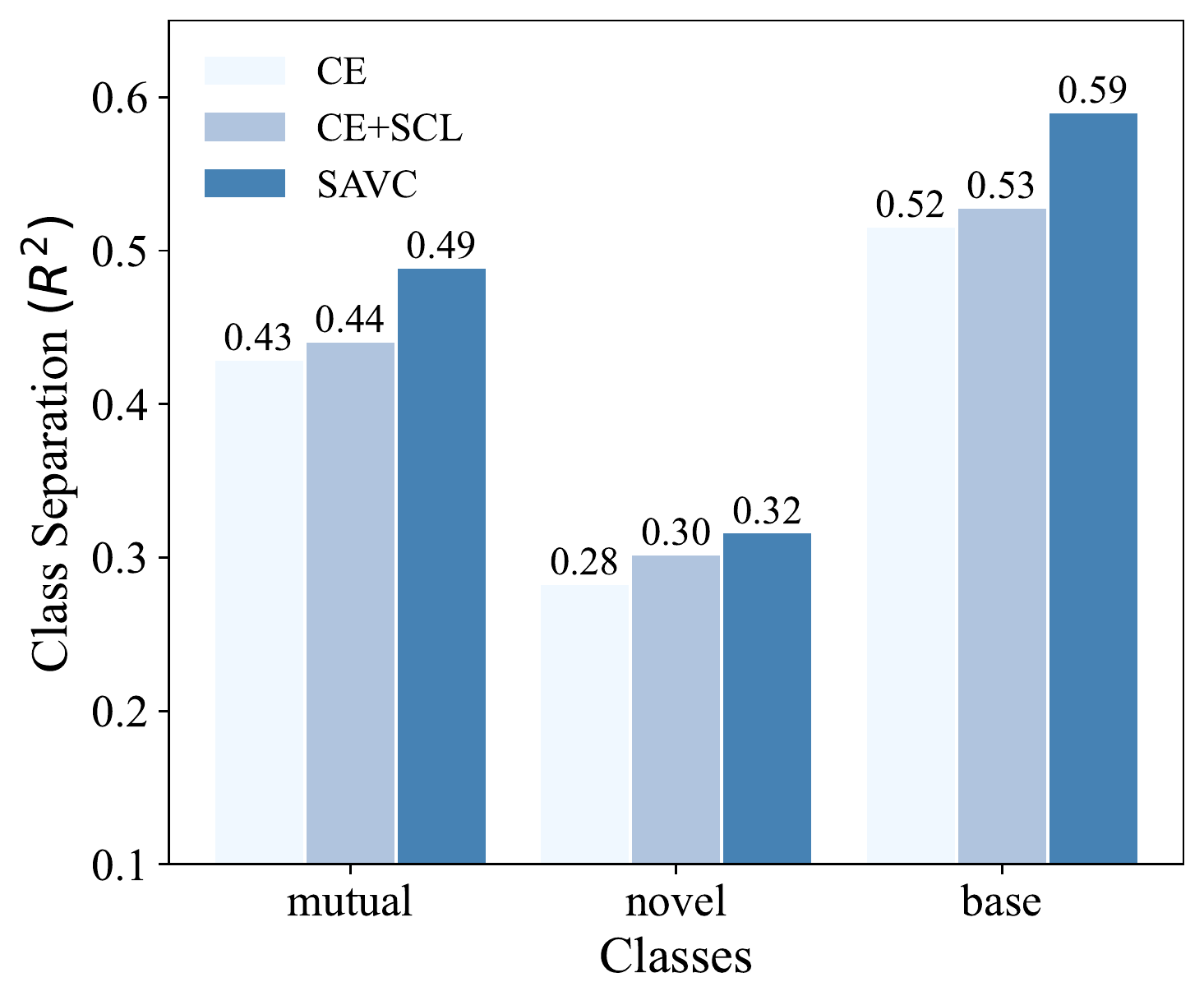}
    \caption{Base, novel and mutual class separation degree.}
    \label{fig:classsep}
  \end{subfigure}
\hfill
  \begin{subfigure}{0.49\linewidth}
    \includegraphics[width=1.05\linewidth]{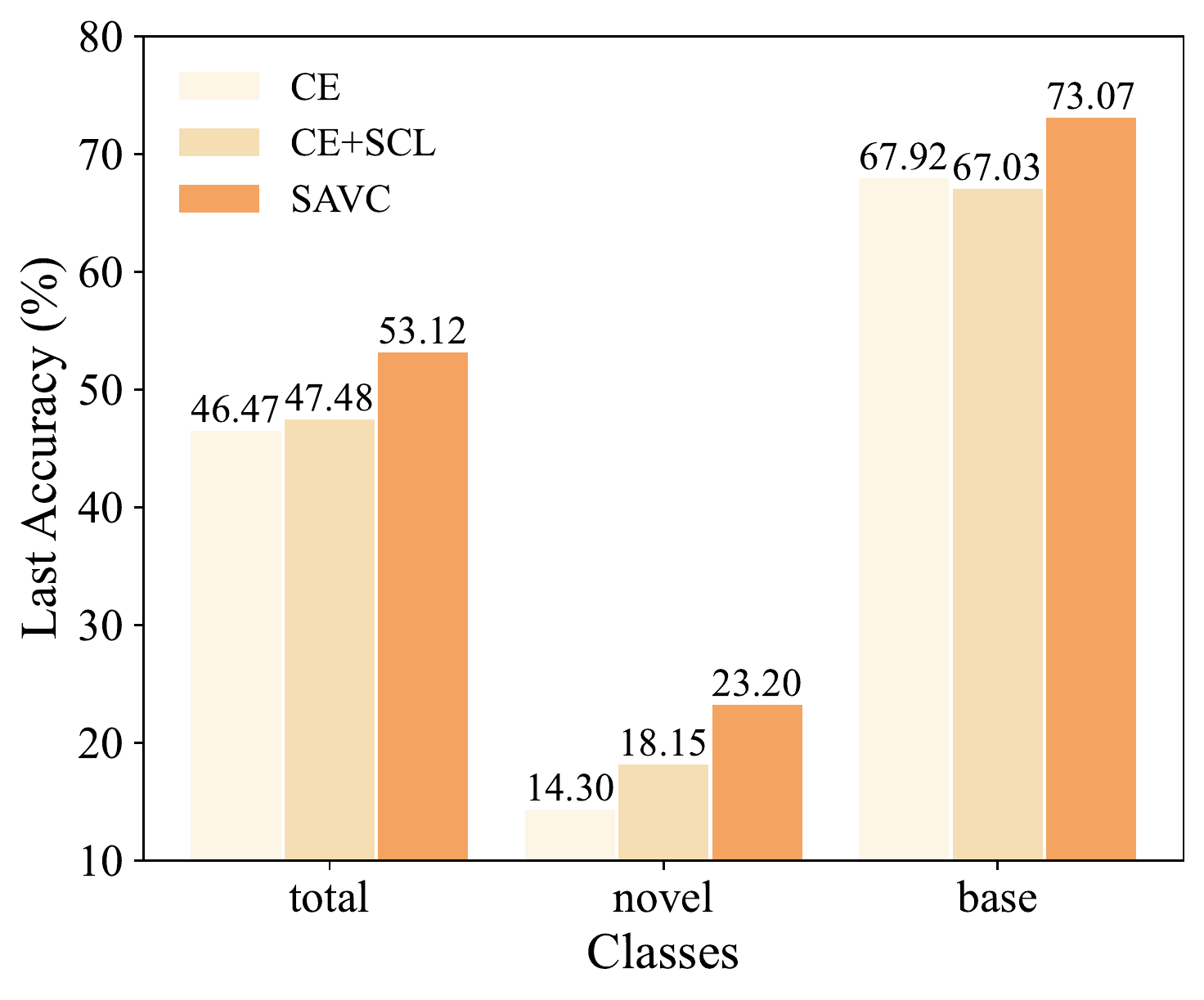}
    \caption{Base, novel and total classification accuracy.}
    \label{fig:classacc}
  \end{subfigure}
  \caption{The relationship between base class separation and novel class generalization. (a) Inter-class distance is defined in ~\cref{eq:inter}. A curve lies more right on the figure denotes greater inter-class distance. (b) Intra-class distance is defined in ~\cref{eq:intra}. A curve lies more left on the figure denotes smaller intra-class distance. (c) Class separation degree ($R^2$) is defined in ~\cref{eq:r^2}.}
  \label{fig:metric}
  \vspace{-10pt}
\end{figure}

\begin{figure*}[!t]
\begin{center}
\includegraphics[width=1.0\textwidth]{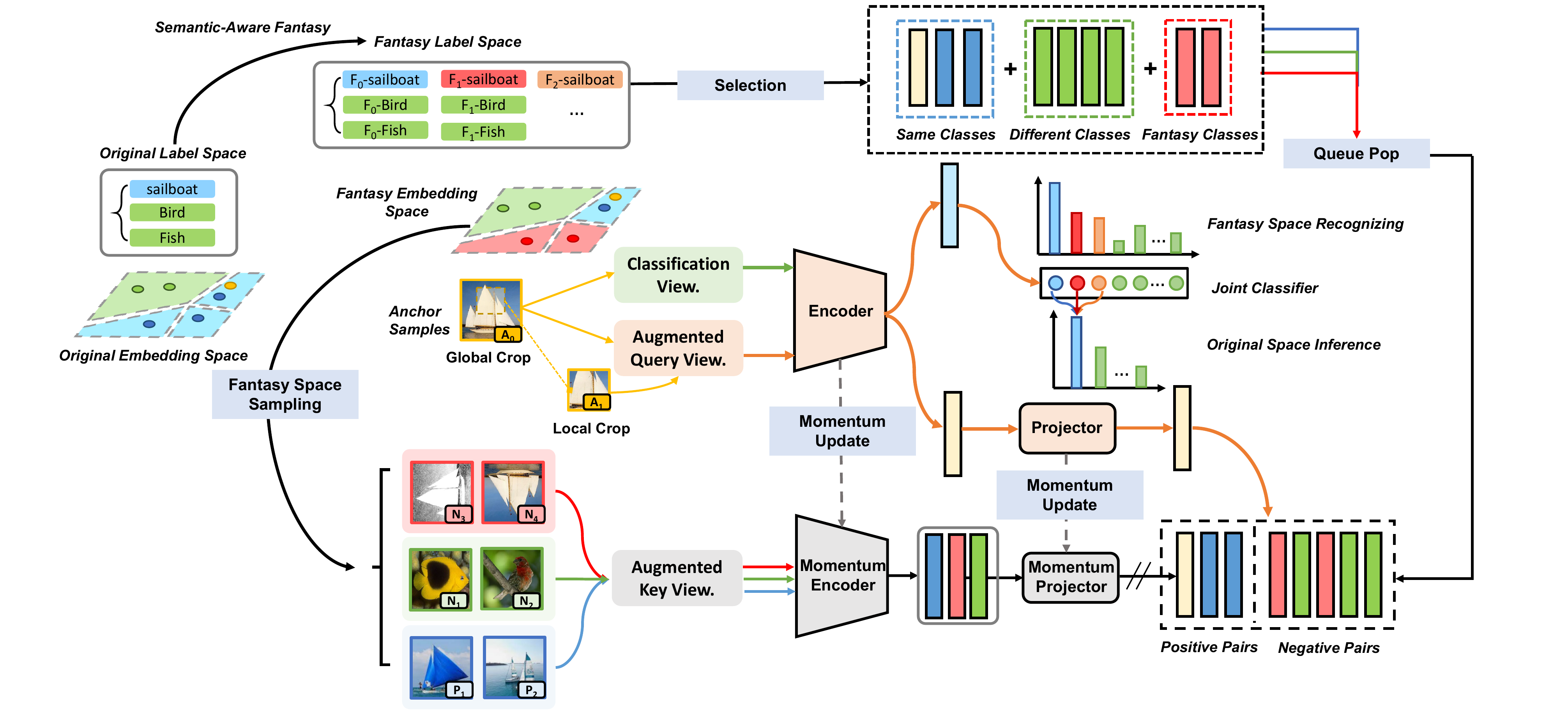}
 \caption{Overall pipeline of our Semantic-Aware Virtual Contrastive (SAVC) framework, which is composed of a classification branch and a contrastive learning branch. The backbone is shared for classification and contrastive query encoding, while the key backbone is updated with a large momentum (`//’ means stop gradient). Both classification and contrastive loss are optimized in fantasy space, while the inference results of different transformations are aggregated and then projected into the original label space.}
 \label{fig:pipeline}
 \end{center}
 \vspace{-15pt}
\end{figure*}

Several provident works~\cite{zhang2021few,shi2021overcoming} recommend an incremental-frozen framework that leverages sufficient data in base session to train the model by optimizing the per-sample loss:
\begin{equation}\label{eq:cls}
{\mathcal{L}_{cls}\left( \phi;\bm{{\rm x}},y \right) = \mathcal{L}}_{ce}\left( {\phi\left( \bm{{\rm x}} \right),y} \right),
\end{equation}
where $\mathcal{L}_{ce}\left( {\cdot,\cdot} \right)$ is cross-entropy loss (CE), $\phi\left( \bm{{\rm x}} \right)$ is the model which can be decomposed into feature extractor and classifier: $\phi(\bm{{\rm x}}) = W^{T}f(\bm{{\rm x}})$, where $\phi(\bm{{\rm x}}) \in \mathbb{R}^{\left| \mathcal{C}^{0} \right| \times 1}$, $W \in \mathbb{R}^{d \times \left| \mathcal{C}^{0} \right|}$ and $f(\bm{{\rm x}}) \in \mathbb{R}^{d \times 1}$. 

After the base training stage, the backbone will be completely fixed, and the classifier is extended in each incremental session: $W = \{ {\bm{w}_{1}^{0},\bm{w}_{2}^{0},\cdots,\bm{w}_{|\mathcal{C}^{0}|}^{0}} \} \cup \cdots \cup \{ {\bm{w}_{1}^{t},\cdots,\bm{w}_{|\mathcal{C}^{t}|}^{t}} \}$. The classifier weight is parameterized by the average embeddings of each class (\ie, prototype), represented as $\bm{w}_{c}^{t} = \frac{1}{n_{c}^{t}}{\sum_{i = 1}^{n_{c}^{t}}{f\left( \bm{{\rm x}}_{c,i} \right)}}$.

In each session, the inference is conducted by the nearest class mean (NCM) algorithm~\cite{mensink2013distance} to evaluate accuracy of all encountered classes. Specifically, we measure the distance between $f(\bm{{\rm x}})$ and all prototypes in classifiers:
\begin{equation}\label{eq:infer}
c_{\bm{{\rm x}}} = \arg{\max\limits_{c,t}{~{\rm sim}\left( {f(\bm{{\rm x}}),\bm{w}_{c}^{t}} \right)}},
\end{equation}
where ${\rm sim}\left( {\bm{{\rm x}},\bm{{\rm y}}} \right) = {{\bm{{\rm x}}^{\mathsf{T}}\bm{{\rm y}}}/\left( {\left\| \bm{{\rm x}} \right\|\left\| \bm{{\rm y}} \right\|} \right)}$ is cosine similarity between two vectors.

\subsection{Base Class Separation Facilitates Novel Class Generalization}
\label{sec:classsep}
The incremental-frozen baseline outperforms a lot of incremental trainable methods~\cite{shi2021overcoming}, demonstrating that preserving old knowledge may matter more than accommodating new knowledge in FSCIL. The guarantee of base performance by the freezing strategy motivates our consideration of what our model should learn in the base training stage for future few-shot generalization.    

An intuitive hypothesis is that, \emph{base class separation facilitates novel class generalization}. If all prototypes are far away from each other, and the features of the same class concentrate around their prototypes, there would be enough place for future class incorporation. Please note that the novel class generalization contains two meanings: without overlapping among novel class sets and with base class embeddings. Previous work demonstrated CE always cannot generalize well due to the poor class margins~\cite{elsayed2018large,liu2016large} and may not be the optimal strategy for FSCIL, which inspires our attempt of supervised contrastive learning (SCL)~\cite{khosla2020supervised}, whose clustering effect has been demonstrated~\cite{zhang2021unleashing}.

We adopt MoCo~\cite{he2020momentum,chen2020improved} framework to implement SCL in our method, for the sake of lower memory budget and storing slowly evolving features of recent batches. For an given image-label pair $\left( {\bm{{\rm x}},y} \right)$, we generate a query view $\bm{{\rm x}}_{q} = {\rm Aug}_{q}(\bm{{\rm x}})$ and a key view $\bm{{\rm x}}_{k} = {\rm Aug}_{k}(\bm{{\rm x}})$ by random augmentations. Then the two augmented images are fed into the query network $g_{q}$ and key network $g_{k}$, getting two $L_{2}$-normalized representations $\bm{q}$ and $\bm{k}$, where $g=h \circ f$ is composed of a feature extractor $f$ and a projector $h$. The two networks share the same architectures, while the key network updates itself with a large momentum. We maintain a feature queue and a label queue with the same length, storing the most current key embeddings and corresponding labels. To this end, given an image $\bm{{\rm x}}$, we have the following per-sample supervised contrastive loss:
\begin{equation}\label{eq:cont}
\begin{split}
&\mathcal{L}_{cont}\left( {g;\bm{{\rm x}},\tau,A} \right) \\
&= - \frac{1}{\left| {P(\bm{{\rm x}})} \right|}{\sum_{\bm{k}_{+} \in P{(\bm{{\rm x}})}}{\log\frac{\exp\left( {{\bm{q}^{\mathsf{T}}\bm{k}_{+}}/\tau} \right)}{\sum\limits_{\bm{k}^{'} \in A{(\bm{{\rm x}})}}{\exp\left( {{\bm{q}^{\mathsf{T}}\bm{k}^{'}}/\tau} \right)}}}},
\end{split}
\end{equation}
where $A(\bm{{\rm x}}) = \bm{k} \cup Q$ is the concatenation of the key embedding $\bm{k}$ of $\bm{{\rm x}}$ and feature queue $Q$, $P(\bm{{\rm x}})$ denotes positive set, ~\ie, those in $A(\bm{{\rm x}})$ belonging to the same class with $\bm{{\rm x}}$. 

In our practice, the feature extractor for query encoding and classification is shared. Put the training of the classification branch and contrastive learning branch together, we form a unified end-to-end framework, and the overall training objective can be concluded as:
\begin{equation}\label{eq:total}
\mathcal{L} = \mathcal{L}_{cls} + \alpha\mathcal{L}_{cont}.
\end{equation}
We would like to refer the model trained with \cref{eq:cls} as CE, and the model trained with \cref{eq:total} as CE+SCL.

We then introduce and define some metrics to measure class separation and generalization, including:

\textbf{1) Inter-class distance.} Given two prototypes $\bm{p}_{m}$ and $\bm{p}_{k}$ of class $m$ and $k$, we compute their distance and get ${\left| \mathcal{C}^{0} \right|\left( {\left| \mathcal{C}^{0} \right| - 1} \right)}/2$ inter-class distances in base session.
\begin{equation}\label{eq:inter}
d_{inter}^{m,k} = 1 - {\rm sim}\left( {\bm{p}_{m},\bm{p}_{k}} \right).
\end{equation}

\textbf{2) Intra-class distance.} Given a class $k$, we compute the average Euclidean distance between all test samples belonging to class $k$ and prototype $\bm{p}_{k}$ to measure the discrepancy within this class. For ${\left| \mathcal{C}^{0} \right|}$ classes in base session, we have:
\begin{equation}\label{eq:intra}
d_{intra}^{k} = 1 - \frac{1}{n_{k}}{\sum\nolimits_{i = 1}^{n_{k}}{{\rm sim}\left( {f\left( \bm{{\rm x}}_{k,i} \right),\bm{p}_{k}} \right)}}.
\end{equation}

\textbf{3) Class Separation ($R^2$).} On the one hand, we refer to Eq. (11) in~\cite{kornblith2021better} to measure the base ($K=\left| \mathcal{C}^{0} \right|$) and novel ($K=\sum_{t = 1}^{T}\left| \mathcal{C}^{t} \right|$) class separation degree respectively:
\begin{equation}\label{eq:r^2}
\begin{split}
    &R^{2} = 1 - {\bar{d}_{within}/\bar{d}_{total}},\\
    &{\bar{d}}_{within} = {\sum\limits_{k = 1}^{K}{\sum\limits_{i = 1}^{n_{k}}{\sum\limits_{j = 1}^{n_{k}}\frac{1 - {\rm sim}\left( {f\left( \bm{{\rm x}}_{k,i} \right),f\left( \bm{{\rm x}}_{k,j} \right)} \right)}{Kn_{k}^{2}}}}},\\
    &\bar{d}_{total} = {\sum\limits_{k = 1}^{K}{\sum\limits_{m = 1}^{K}{\sum\limits_{i = 1}^{n_{j}}{\sum\limits_{j = 1}^{n_{k}}\frac{1 - {\rm sim}\left( {f\left( \bm{{\rm x}}_{k,i} \right),f\left( \bm{{\rm x}}_{m,j} \right)} \right)}{K^{2}n_{k}n_{j}}}}}}.
\end{split}
\end{equation}
On the other hand, we propose a novel metric to evaluate the mutual separation degree between novel and base classes, which is referred to as `mutual class separation ($R_{mul}^{2}$)'. We only need to let $K=\sum_{t = 0}^{T}\left| \mathcal{C}^{t} \right|$ in $\bar{d}_{within}$, and $K=\left| \mathcal{C}^{0} \right|$ and $\sum_{t = 1}^{T}\left| \mathcal{C}^{t} \right|$ respectively in $\bar{d}_{total}$, to compute $R_{mul}^{2}$.

Greater $d_{inter}$ and $R_{base}^2$, smaller $d_{intra}$ represents more compact the base embedding space is. Under the incremental-frozen framework, we compare the result of CE and CE+SCL in \cref{fig:metric} on CIFAR100 benchmark, while $d_{inter}$ and $d_{intra}$ are shown in the cumulative distribution function (CDF) style. We can observe that although CE+SCL gets greater class separation degree and classification accuracy than CE on all classes, it implicitly decreases inter-class distance while decreasing intra-class distance (the CE+SCL curve lies on the left of the CE curve both in \cref{fig:inter} and \cref{fig:intra}). Especially, SCL only provides minor improvement of the mutual class separation $R_{mul}^{2}$, which implies that SCL cannot separate the novel representations from the old distributions (\ie, poor novel class generalization), and is also far from perfect for our objective (see \cref{fig:classsep} and \cref{fig:classacc}).

\subsection{Semantic-Aware Contrastive Class Fantasy}
\label{sec:semantic}
The unsatisfactory performance of SCL, especially the inappreciable improvements of mutual class separation may convey that the semantic information base session provides is not diversified enough, which makes the learned decision boundary very sharp and close to data, resulting in poor base class separation and novel class generalization. Although the rests contained in new classes are invisible in base session, a provident model may be self-sufficient: imagining what would happen in the future and making adequate preparations for it. In other words, our model should use its fantasy to generate many virtual classes with various semantics and fill the unallocated embedding space, all by itself. These virtual classes act as not only class placeholders which preserve enough space for future updates, but also semantic knowledge providers which encourage extensive learning of different semantics for better generalization.

\textbf{Semantic-aware class fantasy.} In our semantic-aware contrastive class fantasy (SAVC) framework, we take inspiration from ~\cite{lee2020self} and generate virtual classes by applying pre-defined transformations in self-supervised learning (SSL)~\cite{zhang2016colorful,pathak2016context,doersch2015unsupervised,gidaris2018unsupervised}. In detail, we first define a discrete transformation (\ie, fantasy) set $\mathcal{F}$ and supposing there are $M$ elements in it,~\ie, the number of transformations. Then, we can generate $M$ transformed samples for an image-label pair $\left( {\bm{{\rm x}},~y} \right)$, denoting $\mathcal{F}{(\bm{{\rm x}},~y)}=\left\{ \left( {\bm{{\rm x}}_{m},~y_{m}} \right) \right\}_{m = 1}^{M}$, where $y_{m} = y \times M + m$. The subscript of $\left( {\bm{{\rm x}}_{m},~y_{m}} \right)$ denotes $m$-th transformation is applied to $\left( {\bm{{\rm x}},~y} \right)$, while $\left( {\bm{{\rm x}}_{1},~y_{1}} \right)$ represents the original image-label pair. 

In this way, we enlarge our label space by $M$ times, and each virtual class can be regarded as a `fine-grained' class derived from the original one. The number and type of the elements in $\mathcal{F}$ convey not only how compact we expect the embedding space to be, but also how fine the semantic grains we want to acquire. Hence we conduct the classification and contrasting training process in finer gains:
\begin{equation}\label{eq:new}
\begin{split}
&\mathcal{L}_{cls}\left( {\phi;\bm{{\rm x}},y,\mathcal{F}} \right) = \frac{1}{M}{\sum_{i = 1}^{M}{\mathcal{L}_{ce}\left( {\phi\left( \bm{{\rm x}}_{m} \right),y_{m}} \right)}},\\
&\mathcal{L}_{cont}\left( {g;\bm{{\rm x}},\tau,A,\mathcal{F}} \right) = \\
&-\frac{1}{M}{\sum\limits_{m = 1}^{M}\frac{1}{\left| {P\left( \bm{{\rm x}}_{m} \right)} \right|}}{\sum_{\bm{k}_{+} \in P{(\bm{{\rm x}}_{m})}}{\!\!\log\frac{\exp\left( {{\bm{q}^{\mathsf{T}}\bm{k}_{+}}/\tau} \right)}{\sum_{\bm{k}^{'} \in A{(\bm{{\rm x}}_{m})}}{\exp\left( {{\bm{q}^{\mathsf{T}}\bm{k}^{'}}/\tau} \right)}}}}.
\end{split}
\end{equation}

\begin{table*}[!t]
  \centering
  \caption{Comparison with SOTA methods on CUB200 dataset for FSCIL. ${*}$: Performances reported by~\cite{tao2020few}. $\Delta_{\text{last}}$: Relative improvements of the last session compared to the Finetune baseline. More results on other datasets can be found in the appendix.}
    \setlength{\tabcolsep}{2.8mm}
\renewcommand{\arraystretch}{1.0}
\resizebox{0.9\linewidth}{!}{
    \begin{tabular}{ccccccccccccc}
    \toprule
    \multirow{2}{*}{Method} & \multicolumn{11}{c}{Acc. in each session (\%) ↑}                                      & \multirow{2}{*}{$\Delta_{\text{last}}$} \\
\cline{2-12}          & 0     & 1     & 2     & 3     & 4     & 5     & 6     & 7     & 8     & 9     & 10    &  \\
    \hline
    Finetune$^{*}$~\cite{tao2020few} & 68.68  & 43.70  & 25.05  & 17.72  & 18.08  & 16.95  & 15.10  & 10.06  & 8.93  & 8.93  & 8.47  & $-$ \\
    iCaRL$^{*}$~\cite{rebuffi2017icarl} & 68.68  & 52.65  & 48.61  & 44.16  & 36.62  & 29.52  & 27.83  & 26.26  & 24.01  & 23.89  & 21.16  & +12.69 \\
    EEIL$^{*}$~\cite{castro2018end}  & 68.68  & 53.63  & 47.91  & 44.20  & 36.30  & 27.46  & 25.93  & 24.70  & 23.95  & 24.13  & 22.11  & +13.64 \\
    TOPIC$^{*}$~\cite{tao2020few} & 68.68  & 62.49  & 54.81  & 49.99  & 45.25  & 41.40  & 38.35  & 35.36  & 32.22  & 28.31  & 26.28  & +17.81 \\
    Rebalancing$^{*}$~\cite{hou2019learning} & 68.68  & 57.12  & 44.21  & 28.78  & 26.71  & 25.66  & 24.62  & 21.52  & 20.12  & 20.06  & 19.87  & +11.40 \\
    SPPR~\cite{zhu2021self}  & 68.68  & 61.85  & 57.43  & 52.68  & 50.19  & 46.88  & 44.65  & 43.07  & 40.17  & 39.63  & 37.33  & +28.86 \\
    F2M~\cite{shi2021overcoming} & 81.07 & 78.16 & 75.57 & 72.89 & 70.86 & 68.17 & 67.01 & 65.26 & 63.36 & 61.76 & 60.26  & +51.79\\
    CEC~\cite{zhang2021few}   & 75.85  & 71.94  & 68.50  & 63.50  & 62.43  & 58.27  & 57.73  & 55.81  & 54.83  & 53.52  & 52.28  & +43.81 \\
    FACT~\cite{zhou2022forward}  & 75.90  & 73.23  & 70.84  & 66.13  & 65.56  & 62.15  & 61.74  & 59.83  & 58.41  & 57.89  & 56.94  & +48.47 \\
    \midrule
    \textbf{SAVC (Ours)}  & \textbf{81.85}  & \textbf{77.92}  & \textbf{74.95}  & \textbf{70.21}  & \textbf{69.96}  & \textbf{67.02}  & \textbf{66.16}  & \textbf{65.30}  & \textbf{63.84}  & \textbf{63.15}  & \textbf{62.50}  & \textbf{+54.03} \\
    \bottomrule
    \end{tabular}%
    }
  \label{tab:CUB}%
  \vspace{-10pt}
\end{table*}%

\begin{figure*}[!t]
\begin{center}
\includegraphics[width=0.9\textwidth]{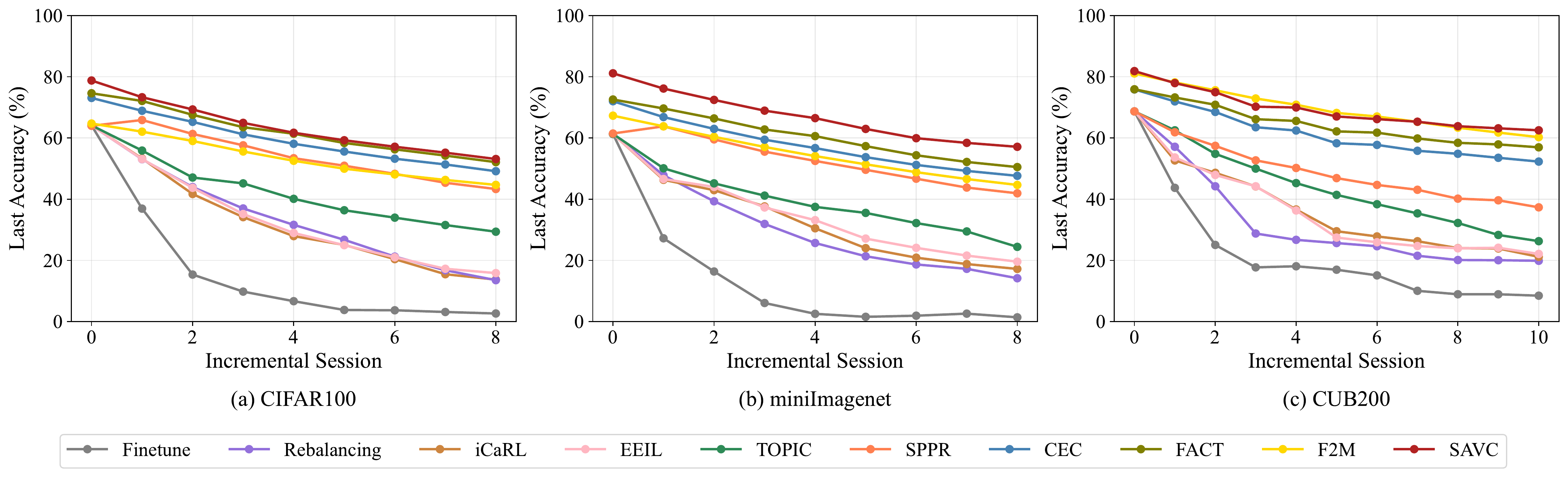}
 \caption{Comparison with SOTA methods on CIFAR100, \emph{mini}ImageNet, and CUB200 benchmarks.}
 \label{fig:acc}
 \end{center}
 \vspace{-15pt}
\end{figure*}

\textbf{Multi-semantic aggregated inference.} With the increase of class, we can acquire various derived class information in finer gains, which enables extending of classifiers with the prototypes of original and pseudo classes: $W = \{ {\bm{w}_{11}^{0},\bm{w}_{12}^{0},\cdots,\bm{w}_{1M}^{0},\cdots,\bm{w}_{|\mathcal{C}^{0}|M}^{0}} \} \cup \cdots \cup \{ {\bm{w}_{11}^{t},\cdots,\bm{w}_{|\mathcal{C}^{t}|M}^{t}} \}$. Prototypes with the same second subscript can naturally form a conditional prototype subset, which denotes a group of clustering centers aware of the same type of fine-grained semantics from the original classes: $W_{m} = \{ {\bm{w}_{1m}^{0},\bm{w}_{2m}^{0},\cdots,\bm{w}_{{|\mathcal{C}^{0}|}m}^{0}} \} \cup \cdots \cup \{ {\bm{w}_{1m}^{t},\cdots,\bm{w}_{{|\mathcal{C}^{t}|}m}^{t}} \}$. In contrast to the traditional single inference in \cref{eq:infer}, we first specify the test sample $\bm{{\rm x}}$ into various fine-grained semantics: $\mathcal{F}{(\bm{{\rm x}})}=\left\{  {\bm{{\rm x}}_{m}}  \right\}_{m = 1}^{M}$. Then we aggregate the inference results from different conditional prototype subsets, which improves accuracy significantly:
\begin{equation}\label{eq:infer_new}
c_{\bm{{\rm x}}} = \arg{\max\limits_{c,t}{{\sum\limits_{m = 1}^{M}{{\rm sim}\left( {f\left( \bm{{\rm x}}_{m} \right),\bm{w}_{cm}^{t}} \right)}}~}}.
\end{equation}
To this end, We compare our SAVC with CE and CE+SCL to verify its effectiveness. As we expected, SAVC effectively increases inter-class distance and decreases intra-class distance in base session, resulting in the best base class separation and novel class generalization (see \cref{fig:metric}).

\begin{table}[!t]
  \centering
  \caption{Statistics of benchmarks datasets. $\mathcal{C}^{base}$: number of classes in base session. $\mathcal{C}^{inc}$: total number of classes in incremental sessions. \#Inc.: number of incremental sessions. Shots: training shots for incremental sessions.}
  \setlength{\tabcolsep}{1.6mm}
\renewcommand{\arraystretch}{0.95}
  \footnotesize
\resizebox{1\linewidth}{!}{
    \begin{tabular}{cccccc}
    \toprule
    Dataset & $\mathcal{C}^{base}$ & $\mathcal{C}^{inc}$ & \#Inc. & Shots & Resolution \\
    \midrule
    CIFAR100~\cite{krizhevsky2009learning} & 60    & 40    & 8     & 5     & 32×32 \\
    \emph{mini}ImageNet~\cite{russakovsky2015imagenet} & 60    & 40    & 8    & 5     & 84×84 \\
    CUB200~\cite{wah2011caltech} & 100   & 100   & 10    & 5     & 224×224 \\
    \bottomrule
    \end{tabular}%
    }
  \label{tab:datasets}%
  \vspace{-13pt}
\end{table}%

\subsection{Discussion}\label{sec:discuss}
Our training pipeline is depicted in \cref{fig:pipeline}. Besides the training regime demonstrated in \cref{sec:semantic}, the proposed SAVC also benefits from the following aspects:

\textbf{Multi-Crop strategy.} Motivated by~\cite{caron2020unsupervised,van2021revisiting}, we add $N$ constraint small crops to capture more local information, which is enforced to overlap with the global crops over a pre-defined threshold ensuring information sharing. These $N$ smaller crops are only served as query views and not fed into the key encoder. In practice, we find optimizing the global and local contrastive loss respectively is helpful, and the overall training objective can be rewritten as:
\begin{equation}\label{eq:total_new}
\mathcal{L} = \mathcal{L}_{cls} + \alpha\mathcal{L}_{cont\_ global} + \beta\mathcal{L}_{cont\_ local}.
\end{equation}

\textbf{Incremental finetuning.} Although there are only a few novel samples available in incremental sessions, it is essential to finetune the pretrained model by \cref{eq:total_new} in each incremental session. Fortunately, we can contrast novel query samples with these slowly evolving key embeddings of base session from the feature queue, which helps push the novel samples away from the distributions of the old classes and benefits novel class generalization.

\section{Experiments}
\label{sec:experiment}

\begin{table*}[htbp]
  \centering
  \caption{Ablation studies on \emph{mini}ImageNet benchmark. \textbf{SCL}, \textbf{F}, \textbf{MC}, \textbf{FT} denote supervised contrastive learning, virtual classes fantasy, multi-crop, incremental finetuning, respectively. $\Delta_{\text{last}}$: Relative improvements of the last sessions compared to the CE baseline.}
  
      \setlength{\tabcolsep}{2.8mm}
\renewcommand{\arraystretch}{1.0}
\resizebox{0.9\linewidth}{!}{
    \begin{tabular}{cccccccccccccc}
    \toprule
    \multirow{2}*{\textbf{SCL}} & \multirow{2}*{\textbf{F}} & \multirow{2}*{\textbf{MC}} & \multirow{2}*{\textbf{FT}} & \multicolumn{9}{c}{Acc. in each session (\%) ↑}                       & \multirow{2}{*}{$\Delta_{\text{last}}$} \\
\cline{5-13}          &       &       &       & 0     & 1     & 2     & 3     & 4     & 5     & 6     & 7     & 8 \\
    \hline
          &       &       &       & 70.27  & 65.17  & 61.13  & 57.80  & 54.80  & 51.84  & 49.14  & 47.03  & 45.54  & $-$\\
    \checkmark     &       &       &       & 72.77  & 67.42  & 63.26  & 59.56  & 56.51  & 53.62  & 50.87  & 48.91  & 47.10  & +1.56\\
    \checkmark      & \checkmark     &      &       & 80.72  & 75.82  & 71.46  & 67.65  & 64.63  & 61.31  & 58.27  & 56.20  & 54.52  & +8.98\\
    \checkmark     & \checkmark     & \checkmark     &       & \textbf{81.12}  & 76.02  & 71.81  & 68.41  & 65.46  & 62.11  & 59.33  & 57.42  & 55.80  & +10.26\\
    \checkmark     & \checkmark     & \checkmark     & \checkmark     & \textbf{81.12}  & \textbf{76.14}  & \textbf{72.43}  & \textbf{68.92}  & \textbf{66.48}  & \textbf{62.95}  & \textbf{59.92}  & \textbf{58.39}  & \textbf{57.11} & \textbf{+11.57}\\
    \bottomrule
    \end{tabular}%
    }
  \label{tab:ablate}%
  \vspace{-12pt}
\end{table*}%

\begin{table}[htbp]
  \centering
  \caption{The concrete meaning of different types of transformation set $\mathcal{F}$. `ALL' denotes using all 0°, 90°, 180°, 270° degrees.}
    \begin{tabular}{cccc}
    \toprule
    $\mathcal{F}$     & Rotation & Color permutation & $M$ \\
    \midrule
    2-fold rorations & 0°, 180° & RGB   & 2 \\
    4-fold rorations & ALL & RGB   & 4 \\
    12-augmentations & ALL & RGB, GBR, BRG & 12 \\
    \bottomrule
    \end{tabular}%
  \label{tab:transform}%
  \vspace{-10pt}
\end{table}%

\subsection{Experimental Setup}
\label{sec:expsettings}
\textbf{Datasets.} Following the benchmark setting~\cite{zhang2021few}, we conduct experiments on three datasets, ~\ie, CIFAR100~\cite{krizhevsky2009learning}, \emph{mini}ImageNet~\cite{russakovsky2015imagenet} and CUB200~\cite{wah2011caltech}, to show the performance of our method. Their statistic characteristics are listed in ~\cref{tab:datasets}.

\textbf{Implementation details.} Following~\cite{zhang2021few}, we adopt ResNet18~\cite{he2016deep} backbones for \emph{mini}ImageNet and CUB200, and ResNet20~\cite{he2016deep} for experiments on CIFAR100. We use SGD with 0.9 momentum to optimize the model. The initial learning rate is 0.1 for CIFAR100 and \emph{mini}ImageNet, and 0.002 for CUB200 in base session. For each incremental learning stage, we only finetune the model for 10 epochs in case of overfitting. We only consider 3 transformation sets $\mathcal{F}$, including 2-fold rotations, 4-fold rotations and 12-augmentations, and their concrete meanings are listed in \cref{tab:transform}. We use 12 augmentations for \emph{mini}ImageNet, and 2-fold rotation for CUB200 and CIFAR100, to fancy virtual classes. We set $\alpha$ as 0.2 and $\beta$ as 0.8 for all datasets to control the impact of global and local contrastive loss.

\subsection{Comparison with State of The Arts}
\label{sec:sotacomp}
We first compare our proposed SAVC with prevailing methods on three FSCIL benchmarks. These methods includes: classical CIL methods,~\ie, iCaRL~\cite{rebuffi2017icarl}, EEIL~\cite{castro2018end}, Rebalancing~\cite{hou2019learning}, incremental-trainable FSCIL methods,~\ie, TOPIC~\cite{tao2020few}, and incremental-frozen FSCIL methods,~\ie, SPPR~\cite{zhu2021self}, 
F2M~\cite{shi2021overcoming},
CEC~\cite{zhang2021few}, and FACT~\cite{zhou2022forward}. We also show a na\"ive baseline that directly finetunes the model with limited data as ‘finetune’. We report the performance curve in \cref{fig:acc} and the detailed values on CUB200 in \cref{tab:CUB}. 

It is observed that ‘finetune’ leaves out preserving old knowledge and gets inferior performance. Classical CIL methods only consider extending the
model with sufficient instances and are not suitable for FSCIL. Incremental-trainable FSCIL methods suffer from overfitting phenomena and perform poorly in incremental sessions. Incremental-frozen FSCIL methods can continually adapt to new classes without any training, which achieve better performance. Under the incremental-frozen framework, our SAVC effectively boosts both base class separation and novel class generalization by recognizing and contrasting semantic-aware virtual classes, consistently achieving the highest average accuracy over all sessions on three datasests.~\eg, our approach gets the final accuracies of $53.12\%$, $62.50\%$ and $57.11\%$ for CIFAR100, CUB200 and \emph{mini}ImageNet, outperforming the current SOTA method FACT by $1.02\%$, $5.56\%$ and $6.62\%$, respectively.

\begin{figure}
  \centering
  \begin{subfigure}{0.46\linewidth}
    \includegraphics[width=1\textwidth]{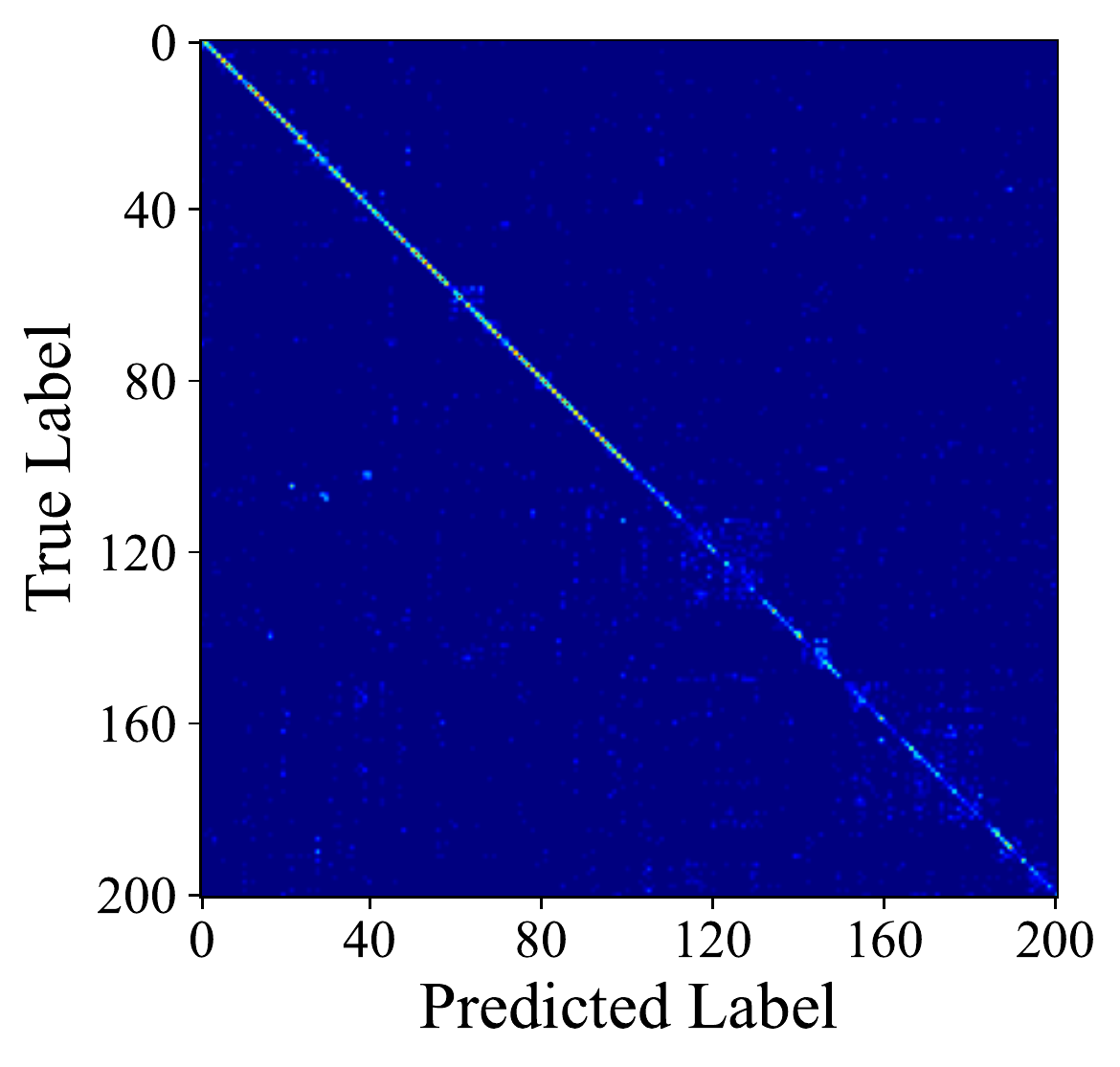}
    \caption{CE.}
    \label{fig:conf_ce}
  \end{subfigure}
  \hfill
  \begin{subfigure}{0.53\linewidth}
    \includegraphics[width=1\textwidth]{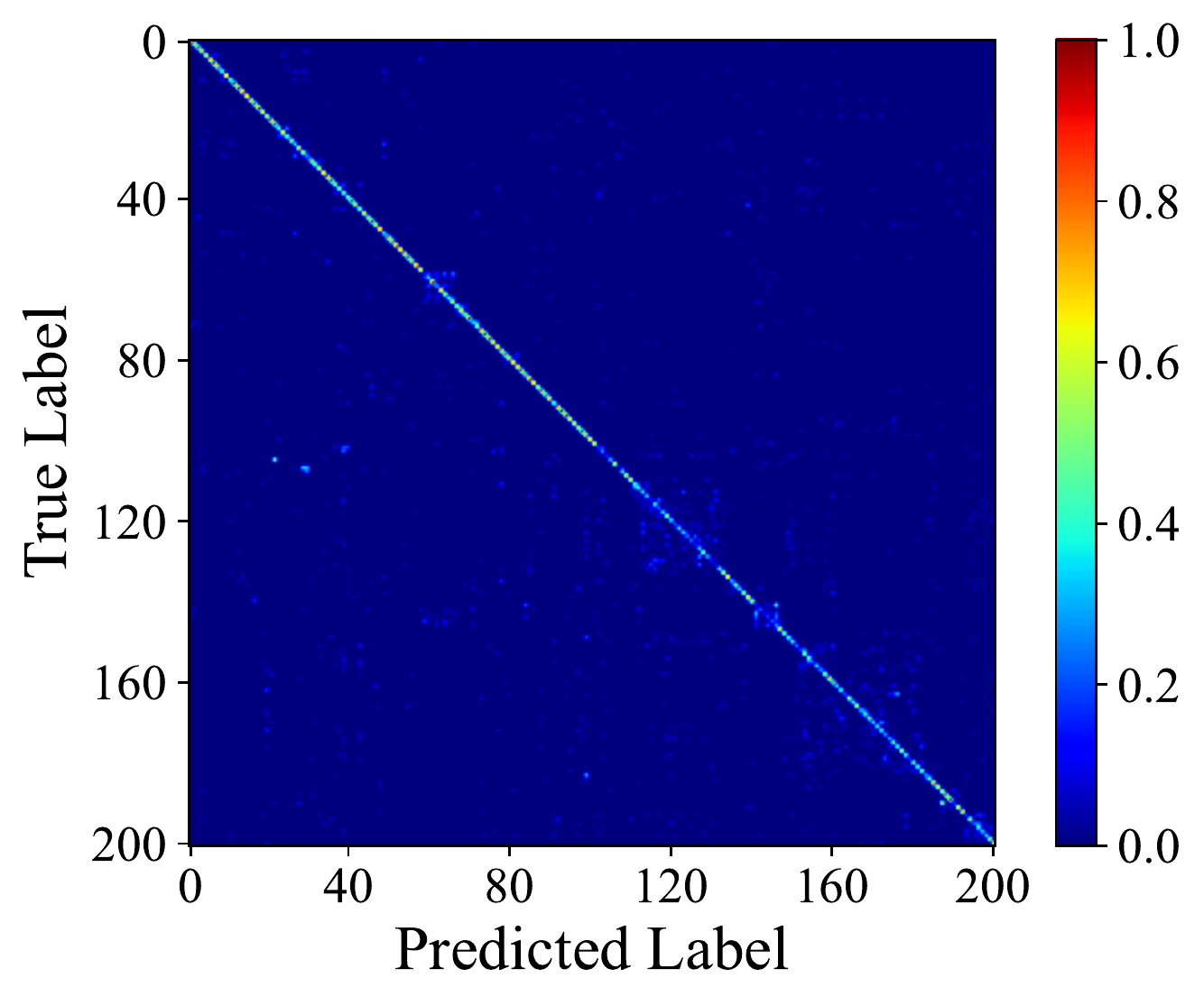}
    \caption{SAVC.}
    \label{fig:conf_savc}
  \end{subfigure}
  \caption{Comparison of the confusion matrices of CE and SAVC methods on CUB200 dataset.}
  \label{fig:confmatrix}
  \vspace{-10pt}
\end{figure}

\subsection{Ablation study}
\label{sec:ablation}
We conduct ablation studies to prove the importance of our proposed components. Built on the incremental-frozen framework, we adopt only CE loss as the baseline, while adding SCL (\textbf{SCL}) and observe the influence this attempt brings (\ie, CE and CE+SCL defined in \cref{sec:classsep}). Then we fantasy (\textbf{F}) contrastive virtual class in CE+SCL and verify its effectiveness. In addition, we consider the influence of the two components mentioned in ~\cref{sec:discuss},~\ie, multi-crop strategy (\textbf{MC}) and finetune (\textbf{FT}) the pre-trained model. We report the results for \emph{mini}ImageNet dataset in \cref{tab:ablate}. 

Compared with the baseline, \textbf{SCL} shows limited influences on improving mutual class separation and lightly improves the performance by 1.56\% in the last session. The fantasy of virtual classes (\textbf{F}) plays a vital part in preventing overlap among old and new distributions, which brings 8.98\% performance improvement. When attending more local information (\textbf{MC}), our model works better,~\ie, 10.26\% gains. Lastly, it is essential to fine-tune the model in incremental session (\textbf{T}), which achieves 11.57\% accuracy gain.

\begin{figure*}[!t]
  \centering
  \begin{subfigure}{0.30\linewidth}
    \includegraphics[width=1.0\linewidth]{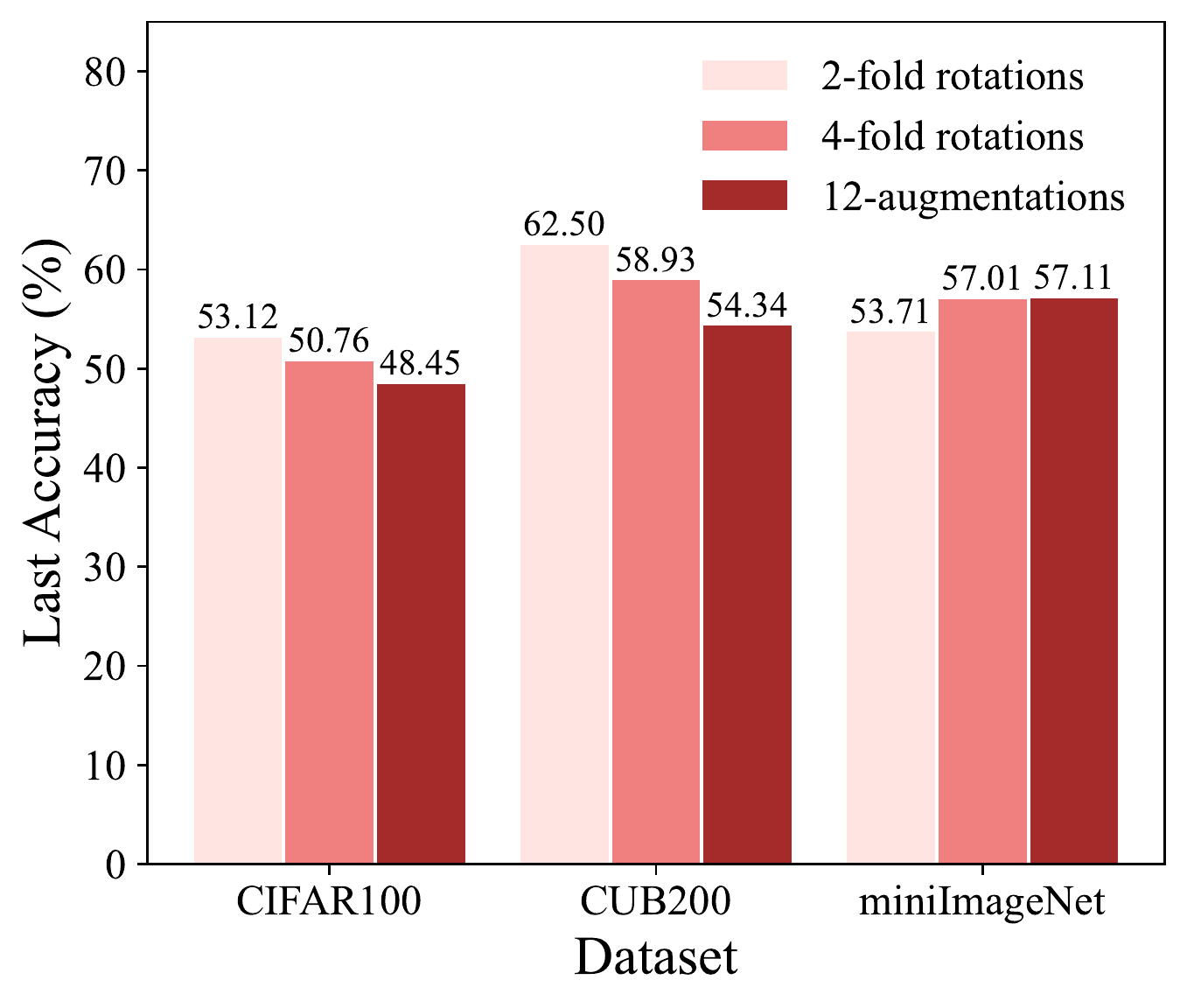}
    \caption{Influence of fantasy methods.}
    \label{fig:fantasy}
  \end{subfigure}
  \hfill
  \begin{subfigure}{0.30\linewidth}
    \includegraphics[width=1.0\linewidth]{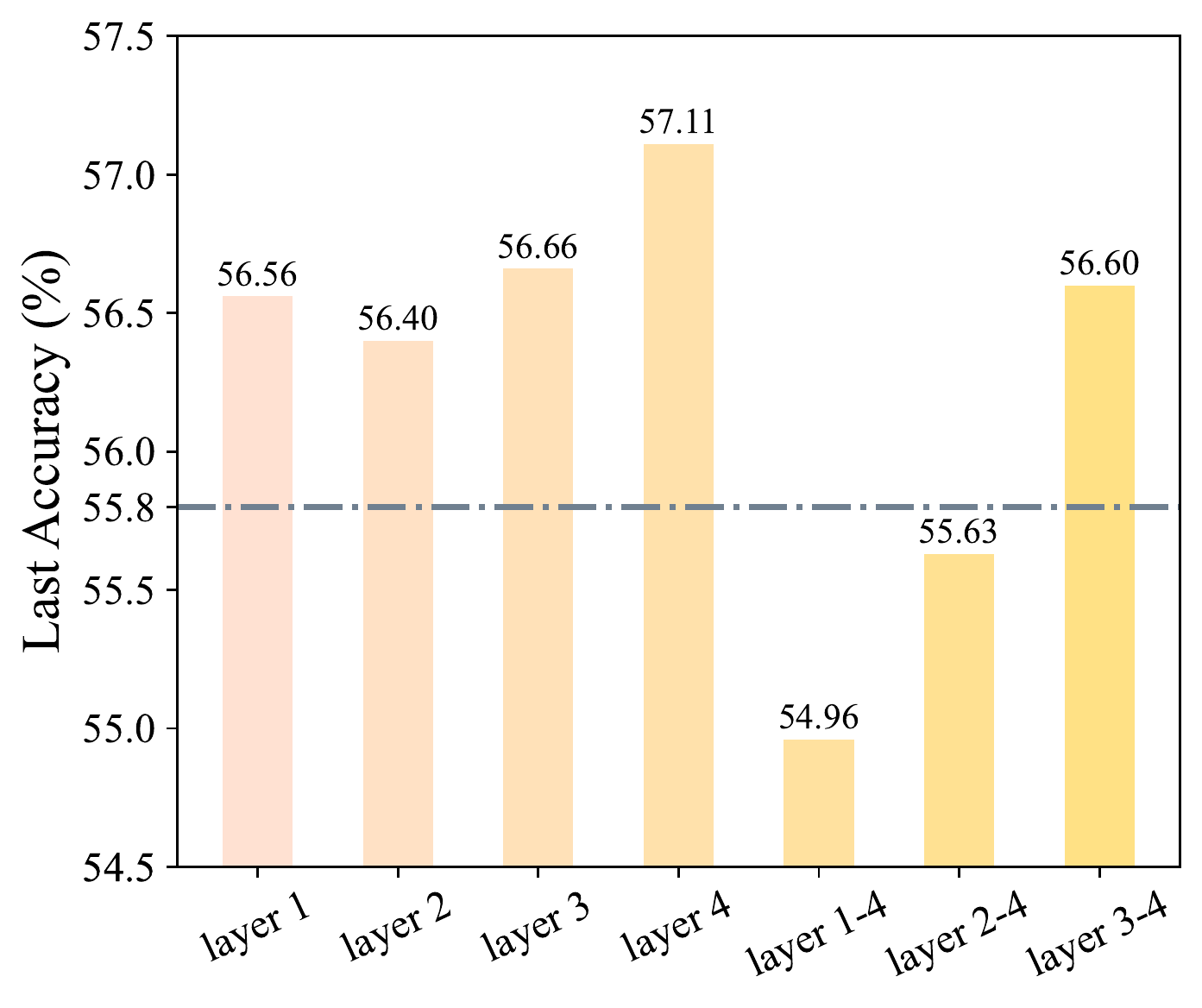}
    \caption{Influence of incremental trainable layers.}
    \label{fig:layers}
  \end{subfigure}
  \hfill
  \begin{subfigure}{0.32\linewidth}
    \includegraphics[width=1.0\linewidth]{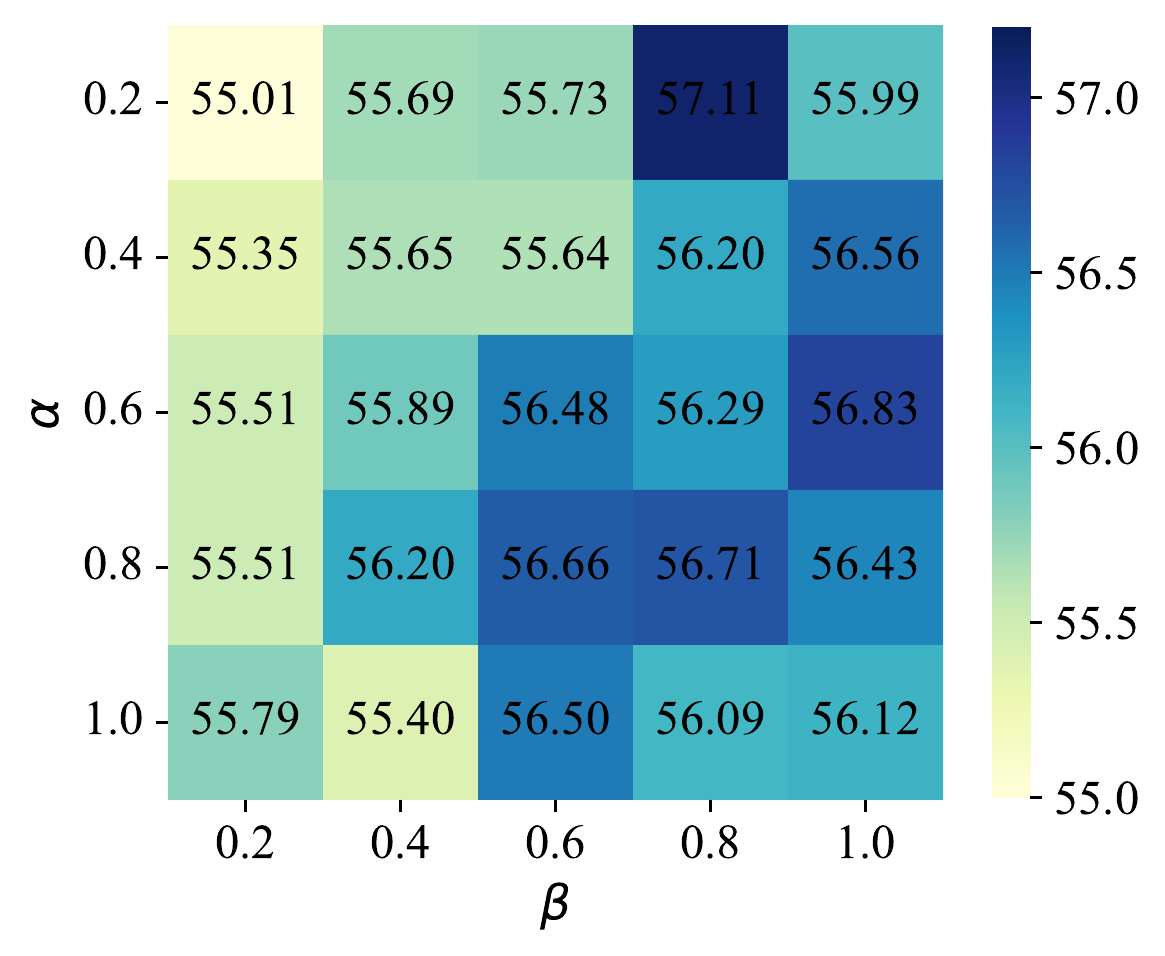}
    \caption{Influence of hyper-parameters.}
    \label{fig:param}
  \end{subfigure}
  \caption{Analysis of detailed hyper configurations of our method: fantasy methods, trainable layers, and hyper-parameters. }
  \label{fig:furanal}
  \vspace{-10pt}
\end{figure*} 

\begin{figure}
  \centering
  \begin{subfigure}{0.32\linewidth}
    \includegraphics[width=1.1\textwidth]{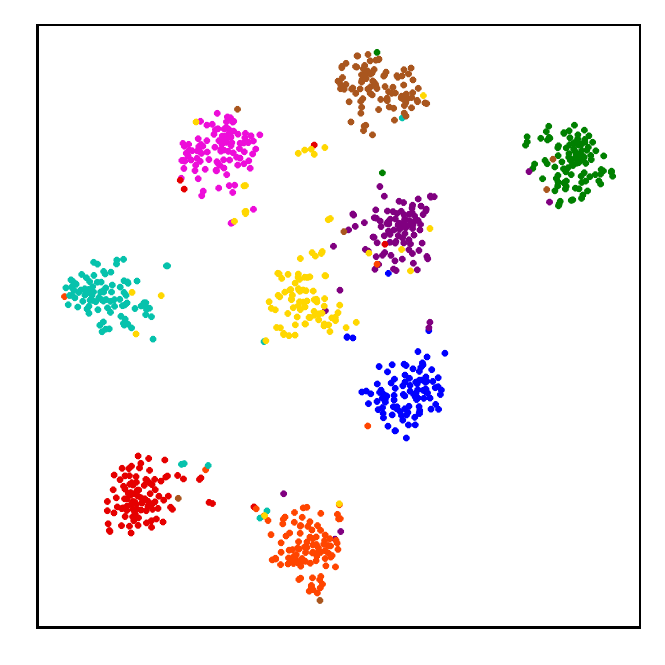}
    \caption{CE.}
    \label{fig:short-a}
  \end{subfigure}
  \hfill
  \begin{subfigure}{0.32\linewidth}
    \includegraphics[width=1.1\textwidth]{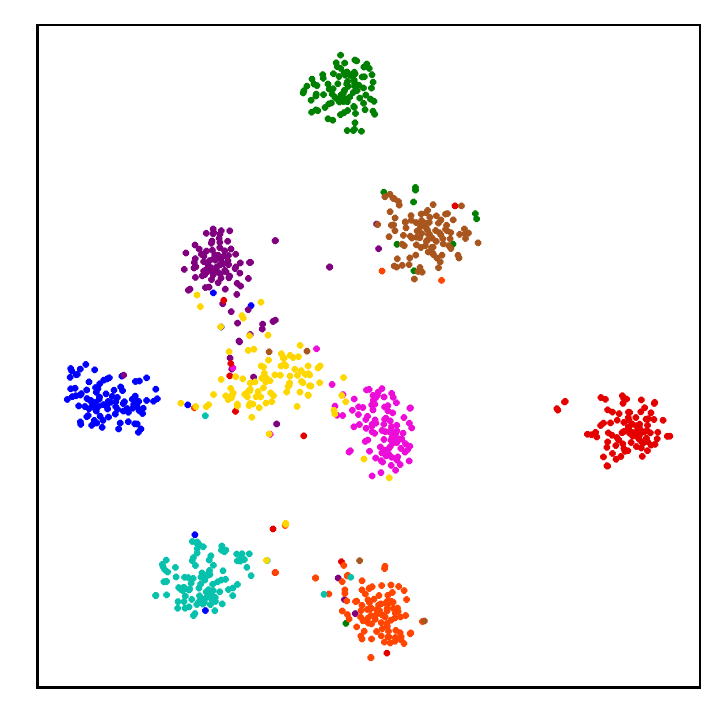}
    \caption{CE+SCL.}
    \label{fig:short-b}
  \end{subfigure}
  \hfill
  \begin{subfigure}{0.32\linewidth}
    \includegraphics[width=1.1\textwidth]{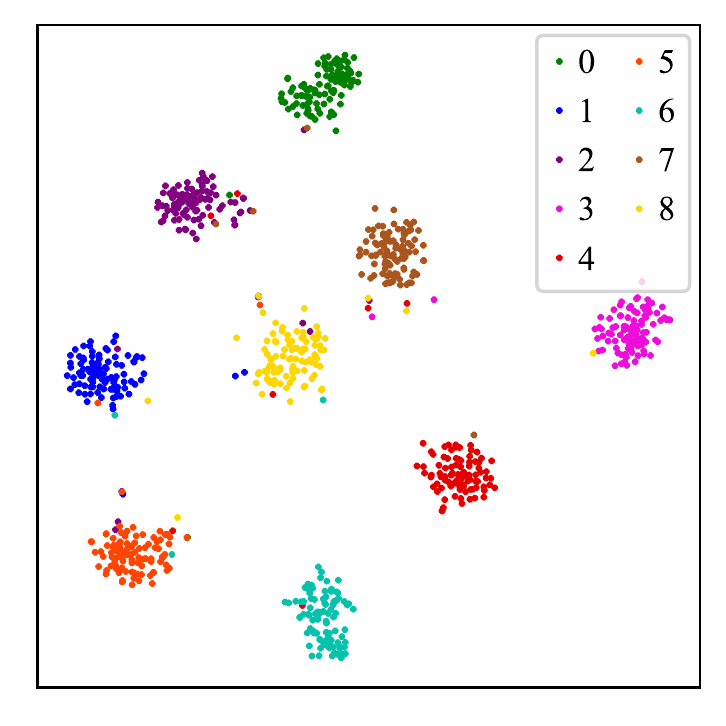}
    \caption{SAVC.}
    \label{fig:short-b}
  \end{subfigure}
  \caption{The t-SNE~\cite{van2008visualizing} visualization on CIFAR100 dataset of the embeddings learned by various methods. Class 0-5 represent the base classes while class 6-8 represent the novel classes. Our SAVC gets the best class separation degree.}
  \label{fig:tsne}
  \vspace{-12pt}
\end{figure}

\subsection{Visualization}
\label{sec:visual}
\textbf{Confusion Matrix.} We visualize the confusion matrix generated by CE and our SAVC on CUB200 dataset after the training of the last session in \cref{fig:confmatrix}. Bright diagonal and dim background indicates higher classification accuracy. It is observed that CE draws a relatively distinct diagonal, but it has poor performance on new classes. In contrast, our method shows better performance on both base and new classes, which proves that our method adapts to novel classes without confusing the previous decision boundaries.

\textbf{Visualization of class separation.} We visualize the embedding space on CIFAR100 dataset with t-SNE~\cite{van2008visualizing} in \cref{fig:tsne}. We randomly choose 6 classes from base classes and 3 classes from novel classes respectively. We can observe that CE has limited clustering effects. Na\"ive SCL slightly improves the class separation degree, but implicitly decreases inter-class distance and remains the mutual class confusion problems unresolved. With the help of virtual classes, Our proposed SAVC can clearly separate base classes and incorporate novel classes with the least overlap.

\subsection{Further Analysis}
\label{sec:furanal}
\textbf{Analysis of fantasy method.} We consider 3 transformation sets $\mathcal{F}$ and show their impacts on our method in \cref{fig:fantasy}. We observe that \emph{mini}ImageNet benefits from complex transformation more, while it would hurt the class separation on CUB200 and CIFAR100. We suppose that CUB200 is already a fine-grained dataset, and the learning process would be hindered by overly fine grains. As for CIFAR100, the image resolutions are too small, and it may be hard to make a distinction between the original and virtual classes. We put discussions on more types of fantasy methods in the appendix.

\textbf{Analysis of trainable layers in incremental sessions.} We update the trainable layers in incremental sessions on \emph{mini}ImageNet benchmark. The results are reported in \cref{fig:layers}. The gray horizontal line indicates without incremental finetuning. It is observed that finetuning fewer and deeper layers gets higher accuracy, while finetuning too many layers harms the former knowledge and shows poorer performance. We infer that the low-level characteristics extracted by shallow layers are shared between classes, so we only need to update high layers to adapt to new classes.

\textbf{Analysis of hyper-parameter.} There are two main hyper-parameters in our loss,~\ie,  coefficient $\alpha$ and $\beta$ which control the impact of global and local supervised contrastive loss in ~\cref{eq:total_new}. We report the accuracy in the last session on \emph{mini}ImageNet dataset varying $\alpha$ and $\beta$ in \cref{fig:param}. Our method is robust to the choices of hyper-parameters, and it achieves the best performance when $\alpha=0.2$ and $\beta=0.8$. For convenience, we fix the value of $\alpha$ and $\beta$ on all datasets.

\section{Conclusion}
In this paper, we solve the FSCIL problems by Semantic-Aware Virtual Contrastive framework (SAVC). Based on the mainstream incremental-frozen framework, both CE and SCL fail to separate base classes and new classes well. We propose to introduce more semantic knowledge by imagining virtual classes. These virtual classes not only hold enough feature space for future updates, but enable a multi-semantic aggregated inference effect and significantly improve classification accuracy. By learning to recognize and contrast in the fantasy space, our method can maintain the separation of all classes and achieve SOTA performance on the three popular FSCIL benchmarks.

\section*{Acknowledgments}
This work is supported by grants from the Key-Area Research and Development Program of Guangdong Province (2020B0101380001), the National Key R$\&$D Program of China (2022ZD0118101), the National Natural Science Foundation of China (61825101, 62202010, 62202014), Shenzhen Basic Research Program (JCYJ20220813151736001), China Postdoctoral Science Foundation (2022M710212), and Huawei Technologies Co., Ltd. The computing resources of Pengcheng Cloudbrain are used in this research.

{\small
\bibliographystyle{ieee_fullname}
\bibliography{egbib}
}

\end{document}